\documentclass{article}

 \usepackage[preprint,nonatbib]{neurips_2025}
 \usepackage[utf8]{inputenc}
\usepackage[T1]{fontenc}
\usepackage{CJKutf8}
\usepackage[utf8]{inputenc} 
\usepackage[T1]{fontenc}    
\usepackage{hyperref}       
\usepackage{url}            
\usepackage{booktabs}       
\usepackage{amsfonts}       
\usepackage{nicefrac}       
\usepackage{microtype}      
\usepackage{xcolor}         
\usepackage{url}            
\usepackage{amsfonts}       
\usepackage{nicefrac}       
\usepackage{microtype}      
\usepackage{multirow}
\usepackage{colortbl}
\usepackage{graphicx}
\usepackage{amsmath}
\usepackage{wrapfig}
\usepackage{tabularx}
\usepackage{url}

\usepackage{makecell}
\title{Zero-Shot Chinese Character Recognition with Hierarchical Multi-Granularity Image-Text Aligning}

\author{
  \textbf{Yinglian Zhu}\thanks{Equal contribution}\quad
  \textbf{Haiyang Yu}\footnotemark[1]\quad
  \textbf{Qizao Wang}\footnotemark[1]\quad
  \textbf{Wei Lu}\quad
  \textbf{Xiangyang Xue}\quad
  \textbf{Bin Li\textsuperscript{\dag}}\quad\\
  Shanghai Key Laboratory of Intelligent Information Processing\\
  School of Computer Science, Fudan University\\
  \texttt{\{ylzhu22, qzwang22, luwei22\}@m.fudan.edu.cn}\\
  \texttt{\{hyyu20, libin, xyxue\}@fudan.edu.cn}
}

\begin{document}
\begin{CJK}{UTF8}{gbsn} 
\maketitle

\begin{abstract}
\label{sec:abs}
Chinese Character Recognition (CCR) is a fundamental technology for intelligent document processing. Unlike Latin characters, Chinese characters exhibit unique spatial structures and compositional rules, allowing for the use of fine-grained semantic information in representation. However, existing approaches are usually based on auto-regressive as well as edit distance post-process and typically rely on a single-level character representation. 
In this paper, we propose a Hierarchical Multi-Granularity Image-Text Aligning (Hi-GITA) framework based on a contrastive paradigm. To leverage the abundant fine-grained semantic information of Chinese characters, we propose multi-granularity encoders on both image and text sides.
Specifically, the Image Multi-Granularity Encoder extracts hierarchical image representations from character images, capturing semantic cues from localized strokes to holistic structures. The Text Multi-Granularity Encoder extracts stroke and radical sequence representations at different levels of granularity. To better capture the relationships between strokes and radicals, we introduce Multi-Granularity Fusion Modules on the image and text sides, respectively. Furthermore, to effectively bridge the two modalities, we further introduce a Fine-Grained Decoupled Image-Text Contrastive loss, which aligns image and text representations across multiple granularities.
Extensive experiments demonstrate that our proposed Hi-GITA significantly outperforms existing zero-shot CCR methods. For instance, it brings about 20\% accuracy improvement in handwritten character and radical zero-shot settings.
Code and models will be released soon.
\end{abstract}
 
\section{Introduction}
\label{sec:intro}
Chinese Character Recognition (CCR) has received growing attention in recent years due to its broad applications, such as historical document analysis~\cite{shi2023m5hisdoc,zhang2025megahan97k} and document intelligence~\cite{9533713,tan2022document,yu2023orientation,zu2023towards}. Chinese characters are completely different from Latin characters due to the large number of characters and their extremely unbalanced distribution in reality~\cite{zhang2025megahan97k}, which makes it easy to encounter the zero-shot problem, i.e., the characters of test sets are absent from the training sets. Therefore, although many studies~\cite{Multi-column,he2016deep,huang2017densely} have been devoted to CCR, it still poses challenges in terms of zero-shot settings.

\begin{figure}[t]
  \centering
   \includegraphics[width=\linewidth]{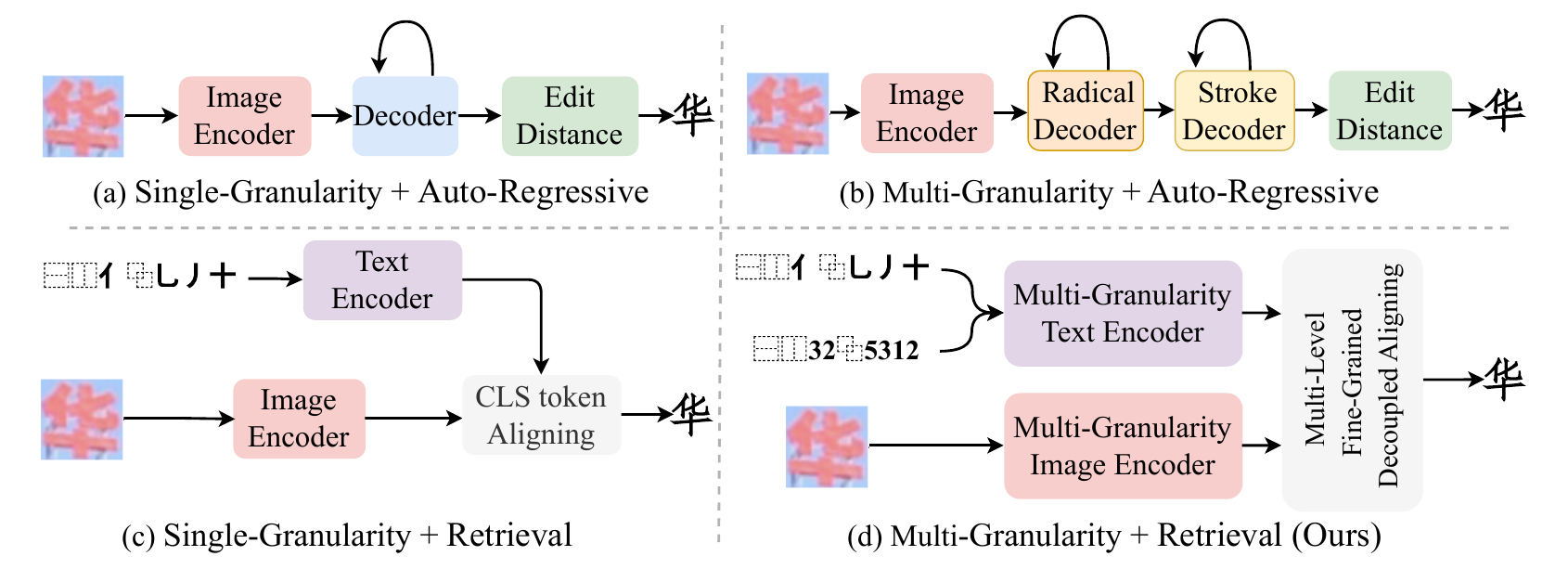}
   \caption{Comparison between different zero-shot CCR approaches. 
   Previous methods are either based on auto-regressive frameworks or only utilized single-granularity representations. We propose a novel retrieval framework using multi-granularity representations and effective image-text aligning.
   }
   \label{fig:intro}
\end{figure}

Distinct from Latin characters, Chinese characters are ideograms with intricate internal spatial structures and can be decomposed into radical or stroke sequences in a specific order (see detailed discussion in Appendix~\ref{appendix:preliminaries}). Therefore, researchers
propose to take advantage of these unique properties to achieve fine-grained representation learning.
Previous radical- and stroke-based approaches~\cite{DenseRAN,chen2022strokebased,chen2021zeroshot} usually address CCR in two stages: predicting stroke/radical sequences and matching them to candidates, as illustrated in Fig.~\ref{fig:intro}(a). 
However, these methods suffer from two major limitations: (1) they fail to leverage the complementary nature of radical- and stroke-level information, and (2) they rely on auto-regressive decoding and edit distance computation, leading to accumulation of errors and burden post-processing.
To address the two limitations, RSST~\cite{yu2023chinese} predicts radical sequences as well as strokes (Fig.~\ref{fig:intro}(b)), while CCR-CLIP~\cite{CCR-CLIP} adopts a CLIP-style~\cite{Clip} architecture to align radical sequences with character images (Fig.~\ref{fig:intro}(c)).

To promote CCR, we propose to address the both limitations in a unified framework.
Given the superior performance and efficiency of CLIP~\cite{Clip} in zero-shot scenarios, we propose a novel CLIP-based framework, named Hi-GITA (Hierarchical Multi-Granularity Image-Text Aligning). 
As shown in Fig.~\ref{fig:intro}(d), it leverages rich multi-granularity representations of Chinese characters on both image and text sides.
Specifically, on the image side, since different image encoder layers characterize learning varying levels of semantics, we guide the encoder to progressively learn stroke, radical, and structure representations. However, radical representations often fail to distinguish stroke differences between visually similar radicals, and stroke representations can benefit from radical-level semantic guidance. Considering the limitations of isolated representations, we introduce the Image Multi-Granularity Fusion Module (MGFM-I) to achieve mutual refinement of strokes and radicals, resulting in refined stroke and refined radical representations.
Accordingly, on the text side, the stroke~\cite{yu2023chinese} and radical~\cite{zhang2018radical} sequences of characters are first encoded by separate encoders. Then, similarly to MGFM-I, to improve the sequence representations of stroke and radical, we design the Text Multi-Granularity Fusion Module (MGFM-T) to achieve symmetric and global interactions between them.

Finally, based on the image and text components across multiple semantic levels (i.e., stroke,
radical, refined stroke, refined radical, structure), we propose a Fine-Grained Decoupled Image-Text Contrastive (FDC) loss to facilitate cross-modal semantic alignment. Since each kind of text sequence is composed of detailed description component and structure component, we decouple each sequence representation and align each separately with its corresponding image counterparts.
Additionally, unlike traditional CLIP-style contrastive losses that only focus on global [CLS] tokens \cite{Clip}, our proposed FDC loss aligns image and text representations across semantic levels in a fine-grained and hierarchical manner.

\noindent\textbf{Our main contributions are summarized as follows:}

\begin{itemize}
   \item  We propose a Hierarchical Multi-Granularity Image-Text Aligning (Hi-GITA) framework based on a contrastive paradigm, which utilizes multi-granularity representations to enhance CCR robustness under the zero-shot settings.
    
   \item  We introduce the Multi-Granularity Fusion Modules on both the image and text sides, enabling effective mutual enhancement between stroke and radical representations.
    
   \item  We design a Fine-Grained Decoupled Image-Text Contrastive loss, which aligns image and text representations across semantic levels in a fine-grained and hierarchical manner.
    
   \item  Extensive experiments show that our proposed Hi-GITA achieves state-of-the-art performance across character and radical zero-shot settings, demonstrating the effectiveness of leveraging the multiple semantic levels of Chinese characters in zero-shot scenarios.
\end{itemize}

\section{Related Work}
\label{sec:related}

Early Chinese Character Recognition (CCR) heavily relies on manual feature designs~\cite{jin2001study,chang2006techniques,su2003novel}. With the development of deep learning, various methods have been developed. In addition, to address the problem of zero shot, radical-based and stroke-based methods have been proposed one after another. In this section, according to the granularity of character splitting, we categorize the existing methods into character-based, radical-based, and stroke-based.

\subsection{Character-based CCR Approach} 
MCDNN~\cite{Multi-column} predicts characters by aggregating features extracted from ResNet-based~\cite{he2016deep} or DenseNet-based~\cite{huang2017densely} networks and achieves human-competitive performance for the first time. Furthermore, the method proposed by~\cite{HWDB} integrates the domain-specific and CNN-based features. Inspired by the human recognition of Chinese characters, DMN~\cite{li2020deep} chooses to match the features of handwritten characters with those of printed characters to accomplish the task of Chinese character recognition. These character-based methods are inherently classification-based methods that are sensitive to variation in the distribution of classes. Therefore, they fail to recognize characters that do not appear in the training set.

\subsection{Radical-based CCR Approach} 
Recently, due to the unique spatial and decomposable characteristics of Chinese characters, many researchers~\cite{DenseRAN,CCR-CLIP,CAO2020107488,LUO2023109598,li2024sidenet,li2025ucr} have begun to use the radical sequence to represent Chinese characters, which reduces the number of final classifications to a certain extent and can alleviate the problem of character zero shot. DenseRAN~\cite{DenseRAN} first models the internal spatial structure of Chinese characters and predicts a sequence of radical representations, which need to be matched with a predefined character decomposition dictionary based on the edit distance to produce the final prediction. HDE~\cite{CAO2020107488} constructs the hierarchical decomposition embedding in the form of a tree and uses embedding matching in the testing phase to make predictions. In a self-information way, CUE~\cite{LUO2023109598} considers the different contributions of radicals in distinguishing characters. CCR-CLIP~\cite{CCR-CLIP} uses CLIP~\cite{Clip} to align the images of the printed characters, while it calculates the similarity between text and image solely based on the similarity score of the global [CLS] token~\cite{dosovitskiy2020image}. Although these methods can alleviate some of the problems of uneven class distribution and character zero-shot, either of them is hard to alleviate the radical zero-shot problem. In other words, radical-based methods cannot well recognize characters with unseen radicals in the test set.

\subsection{Stroke-based CCR Approach} 
Compared with radicals, strokes being the most fundamental components of Chinese characters belong to a much smaller set of categories. As a result, the training set can typically cover all stroke types, eliminating the issue of unseen strokes. This motivates the development of various stroke-based methods for zero-shot Chinese character recognition. The method proposed in~\cite{kim1999decomposition} decomposes each Chinese character into a set of strokes based on the concepts of mathematical morphology. SAE~\cite{chen2022strokebased} adopts a stroke-based autoencoder to model the sophisticated morphology of Chinese characters with a self-supervised method. SD~\cite{chen2021zeroshot} decomposes each character into a sequence of strokes, which are transformed into specific characters by employing a matching-based strategy. 

Other methods~\cite{yu2023chinese, ACPM} propose to use both radical- and stroke-level information to further improve performance, but they are all based on the auto-regressive paradigm, presenting the issues of longer predicted sequences and increased post-processing time. 

\section{Method}
\label{sec:method}

\begin{figure}[tbp]
  \centering
  \includegraphics[width=\linewidth]{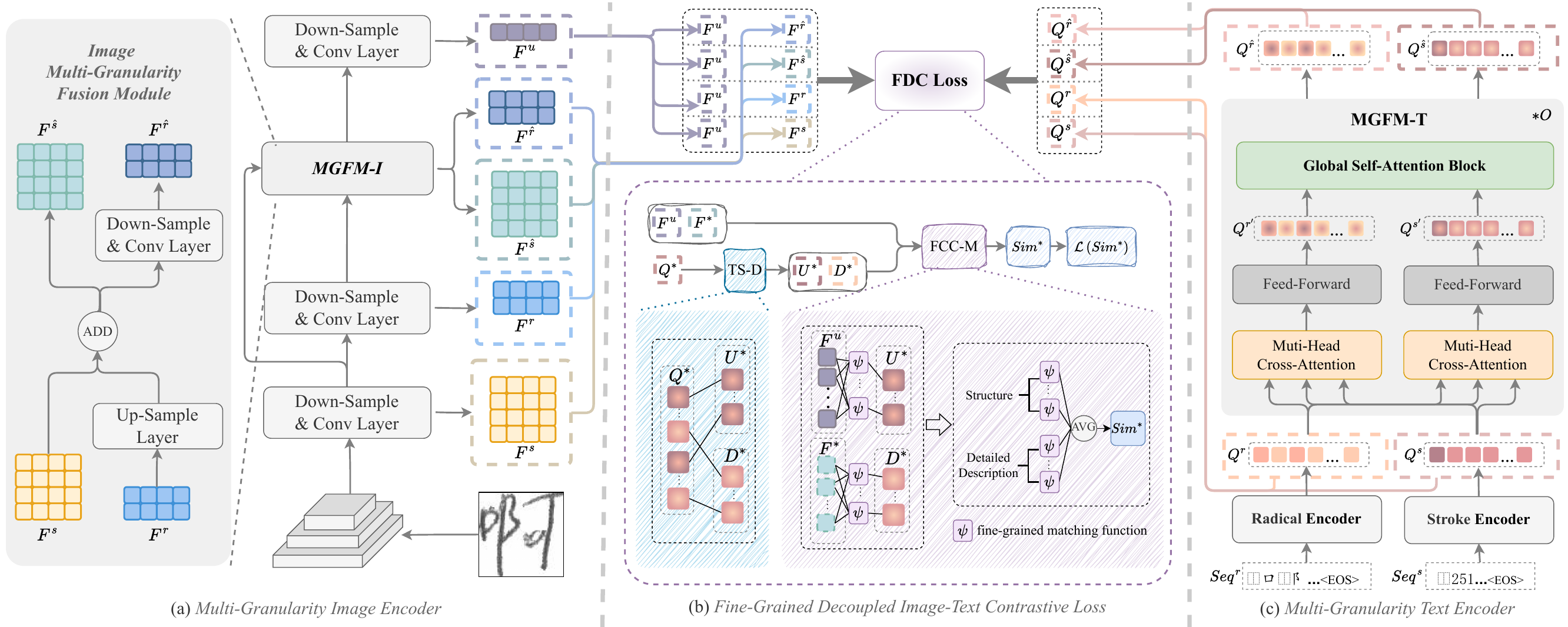}
  \caption{Overview of our proposed framework. 
  On the image side, we learn three levels of representations, i.e., $F^s$, $F^r$, and $F^u$. We also propose MGFM-I to use the stroke and radical representations to refine each other and get their refined representations $F^{\hat{s}}$ and $F^{\hat{r}}$. The text side takes the stroke and radical sequences as inputs, and the stroke and radical encoders separately encode and output stroke and radical sequence representations $Q^s$ and $Q^r$. Then, MGFM-T is applied for their refined representations $Q^{\hat{s}}$ and $Q^{\hat{r}}$. For the Fine-Grained Decoupled Image-Text Contrastive loss, we decouple the text sequence representations and compute the fine-grained image-text contrastive loss at each sematic level.
  }
  \label{fig:overview}
\end{figure}

As shown in Figure~\ref{fig:overview}, our framework consists of three parts: (1) On the image side, we progressively extract hierarchical representations of Chinese characters at three levels, i.e., stroke, radical, and structure.
We also introduce the Image Multi-Granularity Fusion Module (MGFM-I) to achieve a mutual refinement between the stroke and radical representations. 
(2) On the text side, the radical sequence and stroke sequence are processed by radical and stroke encoders, respectively, to extract their sequence representations. Subsequently, in a manner similar to the image side, the Text Multi-Granularity Fusion Module (MGFM-T) integrates the radical and stroke sequence representations to yield their refined counterparts. 
(3) To enable comprehensive cross-modal alignment at different sematic levels, we propose the Fine-Grained Decoupled Image-Text Contrastive (FDC) loss.
\subsection{Multi-Granularity Image Encoder}\label{image_encoder}
As shown in Figure~\ref{fig:overview}(a), given a Chinese character image $x\in\mathbb{R}^{H \times W \times 3}$, where $H$ and $W$ are its height and width, we hierarchically extract three levels of representations from low to high semantics.
Specifically, we employ the first three layers of ResNet-50~\cite{he2016deep} to extract the low-level image representations from the input character image, denoted as $F\in\mathbb{R}^{\frac{H}{4}\times \frac{H}{4} \times D^\prime}$, where $D^\prime$ is the feature dimension. 
Based on $F$, we use two sequential down-sample convolutional layers to obtain the stroke representations $F^{s}\in\mathbb{R}^{\frac{H}{8} \times \frac{W}{8} \times D}$ and the radical representations $F^{r}\in\mathbb{R}^{\frac{H}{16} \times \frac{W}{16} \times D}$, respectively, where $D$ is the feature dimension and the same as that of the text side for alignment. As the initially generated stroke and radical representations are suboptimal, we apply a fusion module (detailed in the following) to sequentially enhance them through mutual interaction. The fusion module takes $F^{s}$ and $F^{r}$ as input and produces the refined stroke and refined radical representations, denoted as $F^{\hat{s}}\in\mathbb{R}^{\frac{H}{8} \times \frac{W}{8} \times D}$ and $F^{\hat{r}} \in\mathbb{R}^{\frac{H}{16} \times \frac{W}{16} \times D}$. Finally, $F^{\hat{r}}$ is processed by a down-sample convolutional layer to obtain the structure representations $F^{u}\in\mathbb{R}^{\frac{H}{32} \times \frac{W}{32} \times D}$.

\textbf{Image Multi-Granularity Fusion Module (MGFM-I).}\label{C2FRL-I}
Since stroke representations are relatively low-level, they can benefit from the semantic guidance of radical representations. 
Moreover, the radical representations only capture the radical category while ignoring differences in strokes, which can easily lead to confusion between visually similar radicals. 
Therefore, in order to refine the stroke representations, we first up-sample the radical representations $F^{r}$ to match the spatial resolution with $F^{s}$. Then, to integrate radical information into the corresponding stroke representations, we adopt element-wise summation for simplicity. Formally,
\begin{equation}
 {F}^{r\prime}=\operatorname{Up-Sample} \left( {F^{r}} \right) ,
\end{equation}
\begin{equation}
 F^{\hat{s}}=F^{s}+{F}^{r\prime},
\end{equation}
where ${F}^{r\prime}\in\mathbb{R}^{\frac{H}{8} \times \frac{W}{8} \times D}$ denotes the upsampled radical representations, ${F}^{\hat{s}}$ denotes the refined stroke representations, 
and $\operatorname{Up-Sample}\left(\cdot\right)$ represents the up-sample layer implemented as one deconvolution layer~\cite{noh2015learning}.

Next, to refine the radical representations $F^{r}$, we follow the reverse process. We apply a down-sample convolutional layer to the refined stroke representations $F^{\hat{s}}$, enabling semantic information to flow from strokes to radicals and yielding refined radical representations $F^{\hat{r}}$.

\subsection{Multi-Granularity Text Encoder}
As shown in Figure~\ref{fig:overview}(c), the radical and stroke encoders first take the radical and stroke sequences as input, respectively. Then, analogous to the image encoder, we propose a multi-granularity fusion module to achieve cross-granularity interaction between strokes and radicals in the text modality.

\textbf{Radical and Stroke Encoders.}
Radical and stroke sequences provide descriptive information about character images at different granularities. Given the two sequences, we employ separate transformer-based encoders to obtain their respective radical and stroke 
sequence representations.
Concretely, we denote the radical and stroke sequences as
$Seq^r = \{ r_1, r_2, \ldots, r_K, \langle\operatorname{eos}\rangle \}$ and
$Seq^s = \{ s_1, s_2, \ldots, s_J, \langle\operatorname{eos}\rangle \}$,
where $K$ and $J$ represent the lengths of the radical and stroke sequences, respectively.
The radical encoder takes $Seq^r$ as input and produces the radical sequence representations
$Q^r = \{ q_1^{r}, q_2^{r}, \ldots, q_K^{r}, q_{\operatorname{eos}}^{r} \}$.
Similarly, the stroke encoder transforms $Seq^s$ into the stroke sequence representations
$Q^s = \{ q_1^{s}, q_2^{s}, \ldots, q_J^{s}, q_{\operatorname{eos}}^{s} \}$.
Both the radical and stroke encoders share the same architecture consisting of $L$ layers of transformer blocks.

\textbf{Text Multi-Granularity Fusion Module (MGFM-T).} 
MGFM-T models the interaction between the stroke and radical sequences. It takes the sequence representations $Q^r\in\mathbb{R}^{\left(K+1\right) \times D}$ and $Q^s\in\mathbb{R}^{\left(J+1\right) \times D}$ as input. 
Firstly, we propose the Symmetrical Cross-Attention Block (SCAB) to mutually enhance the stroke and radical sequence representations through a  symmetrical cross-attention mechanism~\cite{vaswani2017attention}. 
On the one hand, the radical sequence representations $Q^r$ serve as query, while the stroke sequence representations $Q^s$ serve as key and value, allowing the radical sequence representations to attend to stroke semantics. 
On the other hand, the roles are reversed, enabling the stroke sequence representations to attend to radical semantics. Formally,
\begin{equation}
{Q^{s\prime}} =\mathrm{CA}\left(Q^r, Q^s, Q^s\right) ,
\end{equation}
\begin{equation}
{Q^{r\prime}} =\mathrm{CA}\left(Q^s, Q^r, Q^r\right) ),
\end{equation}
where $\mathrm{CA}(\cdot)$ denotes a cross-attention layer followed by a feed-forward layer~\cite{vaswani2017attention}.
To further leverage the complementary semantics of radicals and strokes, we propose the Global Self-Attention Block (GSAB) to refine the obtained radical and stroke sequence representations $Q^{r\prime}$ and $Q^{s\prime}$. 
GSAB applies global semantic interaction between them to get the refined sequence representations $M\in\mathbb{R}^{\left(K+J+2\right) \times D}$, 
which can be expressed by:
\begin{equation}
 M=\operatorname{SA}(\operatorname{Concat}\left(\left[Q^{r\prime}, Q^{s\prime}\right]\right)),
\end{equation}
where $\mathrm{SA}(\cdot)$ denotes a self-attention layer followed by a feed-forward network~\cite{vaswani2017attention}，and $\mathrm{Concat}(\cdot)$ denotes the concatenation operation.

After $O$ layers of sequentially stacked SCAB and GSAB modules, the output first $(K+1)$ tokens are taken as the refined radical sequence representations $Q^{\hat{r}}\in\mathbb{R}^{\left(K+1\right) \times D}$, and the remaining $(J+1)$ tokens are used as the refined stroke sequence representations $Q^{\hat{s}}\in\mathbb{R}^{\left(J+1\right) \times D}$.

\subsection{Fine-Grained Decoupled 
Image-Text Contrastive Loss}\label{loss_function}
With Multi-Granularity Image and Text Encoders, we can extract four types of text sequence representations and five types of image representations.
Each kind of text sequence is composed of a detailed description component and structure component (see detailed discussion in Appendix~\ref{appendix:preliminaries}). Therefore, as illustrated in Figure~\ref{fig:overview}(b), we first decouple the text sequence representations into detailed description component representations and structure component representations. Then, the fine-grained similarity between the image and text component representations is computed using our proposed Fine-Grained Character Components Matching module. Based on this similarity, we apply an improved contrastive image-text loss, which removes duplicate positive examples, to ensure more effective learning.

\textbf{Text Sequence Decoupling (TS-D).}
To better exploit the different roles of detailed description and structure components in the text sequence, we introduce the TS-D module to transform the input text sequence representations into two kinds of component representations: detailed description component representations ${D^{*}}$ and structure component representations ${U^{*}}$, which can be defined as follows:
\begin{equation}
    Q^{*}=D^{*} \cup U^{*},
\end{equation}
where $* \in \{{s, \hat{s}, r, \hat{r}}\}$ and $s$, $\hat{s}$, $r$, and $\hat{r}$ denote the stroke, refined stroke, radical, and refined radical, respectively. $Q^{*}$ denotes different text sequence representations before decoupling.

\textbf{Fine-Grained Character Component Matching (FCC-M).}
As shown in Figure~\ref{fig:overview}(b), after decoupling each of the four sequence representations into detailed description and structure component representations, we match each text component with its corresponding image representations. 
To achieve effective image-text alignment in a contrastive manner, we design a fine-grained matching function $\psi$. It computes the similarities between each text and image token, and then uses the softmax function over the similarities to aggregate the relevance of each text representation across image tokens. $\psi$ is formulated as follows:
\begin{equation}
\label{eq:psi}
    \psi  =  \frac{1}{N}\sum_{n=1}^N \sum_{m=1}^M \alpha_{n,m}\operatorname{sim}^{*}_{n,m},
\end{equation}
\begin{equation}
    \alpha_{n,m} = \frac{\exp\left(\lambda\, \operatorname{Norm} \left(\operatorname{sim}^{*}_{n,m} \right)\right)}{\sum_{m'=1}^M \exp\left(\lambda\, \operatorname{Norm} \left(\operatorname{sim}^{*}_{n,m'} \right) \right)},
\end{equation}
where $*\in \{r, \hat{r}, s,\hat{s}\}$, $\operatorname{sim}_{n,m}$ denotes the inner product between the $n$-th text token and the $m$-th image token, $N$ and $M$ denote the number of text and image tokens, respectively, 
$\lambda$ is a learnable parameter, and $\operatorname{Norm}(\cdot)$ represents L2 normalization.
The scores $\alpha_{n,m}$ serve as attention weights, which are used to reweight the original similarity values $\operatorname{sim}^{*}_{n,m}$. Note that this similarity defines a unidirectional text-to-image matching, as each text token can be aligned with at least one image patch, while not all image patches are associated with a text token.

With the text sequence representations at a specific semantic level (${D}^{*}$ or ${U}^{*}$) along with the corresponding image representations ($F^{*}$ or $F^{u}$), we apply fine-grained matching $\psi$ at the two semantic levels, respectively.
After that, we compute the image-text similarity $Sim^{*}$ as the token-wise average of the fine-grained matching results. 
Formally,
\begin{equation}
Sim^{*}=
\frac{1}{E} \,[\,\underbrace{\psi(F^{*}, D^{*})}_{\substack{\text{Detailed}\\\text{Description}}}
+\underbrace{\psi(F^u, U^{*})}_{\text{Structure}}\,],
\label{eq:sim}
\end{equation}
\begin{equation}
E = 
\begin{cases}
  K + 1, &  * \in \{r,\hat{r}\},\\[4pt]
  J + 1, &  * \in \{s,\hat{s}\}.
\end{cases},
\end{equation}

where $E$ denotes the length of the sequence representation $Q^*$,  
and $\psi(\cdot)$ is the fine-grained matching function, which computes similarities between each text component representations and its corresponding image representations at the same semantic level.
It is worth noting that all text structure component representations $U^{*}$ are aligned with the same image structure representations $F^u$.

\textbf{Improved Image-Text Contrastive Loss.}
\label{loss_fuction}
Unlike general-purpose multimodal datasets, Chinese character recognition datasets involve a relatively limited vocabulary, increasing the likelihood of encountering image-text pairs with repeated characters within the same batch. To address this issue, we exclude image-text pairs that share the same text content as the current pair when computing the contrastive loss.
This strategy ensures that only a single positive example is retained for each contrastive loss computation.
Specifically, at each training iteration, we sample an image-text pair $P_k = (x_k^I, x_k^T)$ from a batch of image-text pairs $\mathcal{B} = \{{(x_i^I, x_i^T)\}}_{i=1}^b$, where $b$ denotes the number of image-text pairs in the batch. We define the set of duplicate positives, i.e., other pair indexes in the batch that share the same text, as $\mathcal{P}=\left\{l \mid x_l^T=x_k^T, l \neq k\right\}$. 
The loss function is as follows:
\begin{equation}
\mathcal{L}\left({Sim}^* \right)   = \frac{1}{2b} \sum_{i=0}^{b} \Biggl[
  \log\Bigl(
    \frac{e^{Sim^*_{i,i}}}
         {\sum_{j\notin \mathcal{P}} e^{Sim^*_{i,j}}}
  \Bigr)
  +
  \log\Bigl(
    \frac{e^{Sim^*_{i,i}}}
         {\sum_{j\prime\notin \mathcal{P}} e^{Sim^*_{j\prime,i}}}
  \Bigr)
\Biggr],
\end{equation}
where $Sim^*_{i,j}$ represents the similarity (following Eq.~\ref{eq:sim}) between the $i$-th text sample and the $j$-th image sample in the batch.

\subsection{Training and Inference}
\label{sec:train_test}
During training, we align image and text representations at five semantic levels: stroke, radical, refined stroke, refined radical, and structure. Since a character can be described by a sequence, we perform image-text alignment separately based on the four types of text sequence representations: stroke sequence, refined stroke sequence, radical sequence, and refined radical sequence.
The overall multi-level FDC loss $\mathcal{L}^{mul}$ is formulated as follows,
\begin{equation}
\mathcal{L}^{mul}=\alpha \left[ \mathcal L \left(Sim^s\right)+ \mathcal L \left(Sim^{\hat{s}}\right) \right] +\beta \left[ \mathcal L \left(Sim^r\right) +  \mathcal L \left(Sim^{\hat{r}}\right) \right],
\label{eq:mul_loss}
\end{equation} 
where $\alpha$ and $\beta$ are hyperparameters used to balance different loss terms.
It is worth noting that with our proposed text sequence decoupling in the FDC loss, the structure component is decoupled. As a result, our method can achieve effective representation learning through multi-level fine-grained decoupled aligning, promoting the generalization and robustness of the model in zero-shot CCR scenarios.

In the inference stage, we use the refined stroke sequence representations $Q^{\hat{s}}$ on the text side, and the refined stroke representations $F^{\hat{s}}$ and structure representations $F^u$ on the image side for retrieval. This design choice stems from the fact that refined strokes provide more fine-grained character representations than refined radicals, while involving a significantly smaller set of categories to avoid stroke zero-shot problems during inference. 
Specifically, we first employ the multi-granularity text encoder to extract the refined stroke sequence representations $Q^{\hat{s}}$ for all candidate characters. Next, the character image is passed through the image encoder to obtain $F^{\hat{s}}$ and $F^u$. Finally, FCC-M is performed to compute the image-text similarity ${Sim}^{\hat{s}}$, based on which the most similar character is identified.
Detailed analysis and experimental results are provided in Section~\ref{Exp_5}.
\section{Experiments}
\label{sec:exp}

\subsection{Experimental Settings}
\label{exp_setting}
\noindent \textbf{Datesets.}
In this paper, we adopt four datasets: HWDB1.0-1.1~\cite{liu2013online}, ICDAR2013~\cite{yin2013icdar}, printed artistic characters~\cite{chen2021zero}, and CTW~\cite{yuan2018chinese}. 

\begin{itemize}
\item \textbf{HWDB1.0-1.1}~\cite{LIU2013155} comprises 2,678,424 images of handwritten Chinese characters, including 3,881 distinct classes, and covers 3,755 commonly-used Level-1 Chinese characters. These images are sourced from a diverse group of 720 writers.
\item \textbf{ICDAR2013}~\cite{yin2013icdar} is collected from 60 writes and contains 224,419 handwritten Chinese character images, covering 3,755 Level-1 characters.
\item \textbf{Printed Artistic Characters}~\cite{chen2021zeroshot} is generated with 105 fonts and contains 394,275 character images, covering 3,755 categories.
\item \textbf{CTW}~\cite{yuan2018chinese} is collected from street view and contains 812,872 images of Chinese characters with 3,650 categories, of which 760,107 images are used for training, and 52,765 images are used for testing. This dataset is very challenging due to its complex background and fonts.
\end{itemize}

\begin{table*}[t]
  \centering
  \small                       
  \setlength{\tabcolsep}{1.7mm}
  \begin{tabular}{l|ccccc|ccccc}
    \toprule
    \multirow{2}{*}{\textbf{Handwritten}} &
      \multicolumn{5}{c|}{$m$ for the Character Zero-Shot Setting} &
      \multicolumn{5}{c}{$n$ for the Radical Zero-Shot Setting} \\
    \cline{2-11}
        & 500 & 1000 & 1500 & 2000 & 2755 & 50 & 40 & 30 & 20 & 10 \\ \hline

    DenseRAN~\cite{DenseRAN}  & 1.70  & 8.44  & 14.71  & 19.51  & 30.68  & 0.21  & 0.29  & 0.25  & 0.42  & 0.69  \\
    HDE~\cite{CAO2020107488}  & 4.90  & 12.77  & 19.25  & 25.13  & 33.49  & 3.26  & 4.29  & 6.33  & 7.64  & 9.33  \\
    SD~\cite{chen2021zeroshot}& 5.60  & 13.85  & 22.88  & 25.73  & 37.91  & 5.28  & 6.87  & 9.02  & 14.67  & 15.83  \\
    CUE~\cite{LUO2023109598}  & 7.43  & 15.75  & 24.01  & 27.04  & 40.55  & - & - & - & - & - \\
    SideNet~\cite{li2024sidenet} & {5.1 } & {16.2 } & {33.8 } & {44.1 } & {50.3 } & - & - & - & - & - \\
    RSST~\cite{yu2023chinese} & 11.56  & 21.83  & 35.32  & 39.22  & 47.44  & 7.94  & 11.56  & 15.13  & 15.92  & {20.21}  \\
    UCR~\cite{li2025ucr} & {15.16} & {24.04 } & {37.46 } & {42.21} & {49.36 } & {10.57} & \underline{14.34
    } & \underline{17.68} & \underline{20.05} & \underline{23.41} \\
    CCR-CLIP~\cite{CCR-CLIP}  & \underline{21.79 } & \underline{42.99 } & \underline{55.86 } & \underline{62.99 } & \underline{72.98 } & \underline{11.15 } & {13.85 } & {16.01 } & {16.76 } & 15.96 \\ \hline
    \rowcolor{gray!30}
    Ours & \textbf{45.34 } & \textbf{66.56 } & \textbf{73.06 } & \textbf{80.88 } & \textbf{85.38 } & \textbf{31.13 } & \textbf{37.88 } & \textbf{44.55 } & \textbf{47.85 } & \textbf{44.18 } \\
    \bottomrule
    
    \toprule
    \multirow{2}{*}{\textbf{Printed Artistic}} &
      \multicolumn{5}{c|}{$m$ for the Character Zero-Shot Setting} &
      \multicolumn{5}{c}{$n$ for the Radical Zero-Shot Setting} \\
    \cline{2-11}
        & 500 & 1000 & 1500 & 2000 & 2755 & 50 & 40 & 30 & 20 & 10 \\ \hline
    DenseRAN~\cite{DenseRAN} & 0.20  & 2.26  & 7.89  & 10.86  & 24.80  & 0.07  & 0.16  & 0.25  & 0.78  & 1.15  \\
    HDE~\cite{CAO2020107488} & 7.48  & 21.13  & 31.75  & 40.43  & 51.41  & 4.85  & 6.27  & 10.02  & 12.75  & 15.25  \\
    SD~\cite{chen2021zeroshot}  & 7.03  & 26.22  & 48.42  & 54.86  & 65.44  & 11.66  & 17.23  & 20.62  & {31.10 } & {35.81 } \\
    SideNet~\cite{li2024sidenet} & {14.6 } & {40.7 } & {60.6 } & \underline{69.8 } & \underline{73.8 } & - & - & - & - & - \\
    RSST~\cite{yu2023chinese} & {23.12 } & {42.21 } & {62.29 } & {66.86 } & {71.32 } & {13.90 } & {19.45 } & {26.59 } & {34.11 } & {38.15 } \\ 
    UCR~\cite{li2025ucr} & \underline{28.52} & \underline{45.04
    } & \underline{63.53 } & {67.23} & {71.55 } & \underline{19.98} & \underline{28.07
    } & \underline{35.42} & \underline{42.33} & \underline{50.45} \\\hline
    \rowcolor{gray!30}
    Ours & \textbf{30.81 } & \textbf{61.31 } & \textbf{81.29 } & \textbf{84.93 } & \textbf{88.22 } & \textbf{34.58 } & \textbf{43.15 } & \textbf{36.61 } & \textbf{58.7 } & \textbf{51.88 } \\
    \bottomrule

    \toprule
    \multirow{2}{*}{\textbf{Scene}} &
      \multicolumn{5}{c|}{$m$ for the Character Zero-Shot Setting} &
      \multicolumn{5}{c}{$n$ for the Radical Zero-Shot Setting} \\
    \cline{2-11}
        & 500 & 1000 & 1500 & 2000 & 3150 & 50 & 40 & 30 & 20 & 10 \\ \hline
    DenseRAN~\cite{DenseRAN}  & 0.15  & 0.54  & 1.60  & 1.95  & 5.39  & 0 & 0 & 0 & 0 & 0.04  \\
    HDE~\cite{CAO2020107488}  & 0.82  & 2.11  & 3.11  & 6.96  & 7.75  & 0.18  & 0.27  & 0.61  & 0.63  & 0.90  \\
    SD~\cite{chen2021zeroshot} & 1.54  & 2.54  & 4.32  & 6.82  & 8.61  & 0.66  & 0.75  & 0.81  & 0.94  & 2.25  \\
    RSST~\cite{yu2023chinese} & 1.41  & 2.53  & 4.59  & 9.32  & 13.02  & {1.21 } & 1.29  & 1.89  & {2.90 } & 3.88  \\
    UCR~\cite{li2025ucr} & {2.05} & {3.48
    } & {5.63 } & {9.44} & {14.13 } & \textbf{1.88} & \underline{2.24    } & \underline{2.93} & \textbf{3.51} & \underline{4.37} \\
    CCR-CLIP~\cite{CCR-CLIP}  & \underline{3.55 } & \underline{7.70 } & \underline{9.48 } & \underline{17.15 } & \underline{24.91 } & 0.95  & {1.77 } & {2.36 } & 2.59  & {4.21 } \\    \hline
    \rowcolor{gray!30}
    Ours & \textbf{3.92 } & \textbf{9.91 } & \textbf{14.98 } & \textbf{19.9 } & \textbf{31.4 } & \underline{1.87 } & \textbf{2.67 } & \textbf{3.34 } & \underline{3.27 } & \textbf{4.83 } \\
    \bottomrule
  \end{tabular}

  \caption{Comparison results on the handwritten dataset, printed artistic dataset, and scene dataset. Results in the character and radical zero-shot settings are reported.}
  \label{tab:Sota_compare}
\end{table*}

\noindent \textbf{Zero-Shot Settings.}
Following~\cite{chen2021zeroshot}, we construct datasets for both the character and radical zero-shot settings. (1) For the character zero-shot setting, we assign samples from the first $m$ classes to the training set and those from the last $k$ classes to the test set, ensuring that the test classes are unseen during training. For the handwritten character dataset, we use HWDB1.0-1.1~\cite{LIU2013155} as the training set with $m \in \{500, 1000, 1500, 2000, 2755\}$ and ICDAR2013~\cite{yin2013icdar} as the test set, and $k$ is fixed at 1000. For the printed artistic character dataset, we follow the same experimental settings as the handwritten dataset, using the same values of $m$ and $k$ for constructing the character zero-shot setting. For the scene character dataset CTW, $m \in \{500, 1000, 1500, 2000, 3150\}$ is used for training, and the remaining $k = 500$ classes are used for testing.
(2) For the radical zero-shot setting, we first compute the frequency of each radical in the lexicon. Characters containing at least one radical that appears fewer than $n$ times are assigned to the test set, while the remaining characters are used for training. We experiment with $n \in \{10, 20, 30, 40, 50\}$.

\noindent \textbf{Evaluation metric.}
Following previous works~\cite{cao2020zero,DenseRAN,xiao2019template,zhang2020radical}, we use Character ACCuracy (CACC) for performance evaluation.

\noindent \textbf{Implementation Details.}
Our method is implemented with PyTorch, and all experiments are conducted on an NVIDIA RTX 4090 GPU with 24 GB of memory. Input images are resized to 128$\times$128.
We adopt Adam optimizer~\cite{kingma2014adam} with an initial learning rate of $1 \times 10^{-4}$. The batch size is set to 80.
The radical encoder and stroke encoder are each composed of $L=3$ layers. Both MGFM-I and MGFM-T contain $O=3$ layers.
The maximum lengths of the stroke and radical sequences are both set to 50. Sequences shorter than the maximum length are padded. 
The hyperparameters $\alpha$ and $\beta$ are set to 1.0 and 0.1.

\subsection{Experimental Results}
\noindent \textbf{Zero-Shot Recognition Results.} 
As shown in the top two sections of Table~\ref{tab:Sota_compare}, on the handwritten dataset and the printed artistic dataset, our method outperforms the previous state-of-the-art models~\cite{CCR-CLIP,li2025ucr} by a large margin. Specifically, in the character and radical zero-shot settings, it achieves average improvements of 18.92\% and 23.79\% on the handwritten dataset, and 15.07\% and 9.73\% on the printed artistic dataset, respectively.
Compared to the handwritten and printed artistic datasets, CTW contains more complex backgrounds. Therefore, as shown in the bottom section of Table~\ref{tab:Sota_compare}, prior methods all show suboptimal performance on this dataset.
Our proposed method achieves improvements of 3.46\% and 0.21\% over the previous state-of-the-art in the character and radical zero-shot settings, respectively. The relatively mild gain is mainly due to the presence of severely blurred images in CTW.

\noindent \textbf{Closed-Set Recognition Results.}
To assess whether our method generalizes beyond the zero-shot scenarios, we also conduct closed-set recognition experiments on the handwritten dataset. Detailed results are provided in Appendix~\ref{appendix:closed-set}.

\subsection{Inference with Different Representations}
\label{Exp_5}
 \begin{figure}[tbp]
  \centering
  \includegraphics[width=\linewidth]{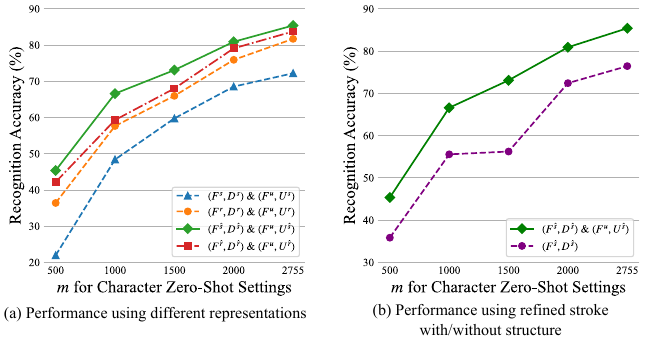}
  \caption{Performance comparison of different representations used for inference. $\left( \cdot\right)$ represents a pair of image representations and their text counterparts used for retrieval. }
  \label{fig:diff-rep}
\end{figure}
Our method jointly optimizes five semantic levels during training, and leverages multiple representations during inference. 
We evaluate the performance of different semantic levels of representations in Figure~\ref{fig:diff-rep}. We first evaluate the performance by matching both detailed description and structure components from four types of text sequence representations with their corresponding image representations. As shown in Figure~\ref{fig:diff-rep}(a), refined stroke with structure representations achieve the best performance, and the performance ranking of different semantic representations (from highest to lowest) is: refined stroke, refined radical, radical, and stroke. 
We also observe that compared to radical representations, stroke representations are relatively low-level and lack architectural information, which limits their ability to capture the overall representation of a character. 
Meanwhile, both refined stroke and refined radical representations outperform their unrefined counterparts. This improvement can be attributed to the incorporation of cross-granularity interaction of MGFM, which facilitates a more robust architectural representation for each character. Refined strokes outperform refined radicals, since the stroke-based representations are better able to capture fine-grained character components after MGFM. 

Then, we evaluate the performance using refined stroke with/without structure for recognition. As shown in Figure~\ref{fig:diff-rep}(b), removing the structure component leads to an average performance drop of 10.97\% in the character zero-shot settings, compared to using both components. Structure helps the model better capture the positional relationships between radicals, thereby enabling the construction of more hierarchical character representations. 
We also evaluate matching with only the structure component. It shows a dramatic performance collapse, with average recognition accuracies dropping to only 1.31\% in the character zero-shot settings. It is reasonable since structural representations primarily encode the positional relationships between radicals, while the main semantic information for character recognition resides in the detailed description component.
More experimental results in the radical zero-shot settings are shown in Appendix~\ref{appendix:RZS-diff_rep}.

\begin{table}[tbp]
  \centering
  \small
  \renewcommand{\arraystretch}{1.0}
  \setlength{\tabcolsep}{1.5mm}
  \begin{tabularx}{\linewidth}{
      >{\centering\arraybackslash}X
      >{\centering\arraybackslash}X
      >{\centering\arraybackslash}X
      >{\centering\arraybackslash}X
      |>{\centering\arraybackslash}X
      |>{\centering\arraybackslash}X
    }
    \toprule
    \textbf{MGFM-T} & \textbf{FCC-M} & \textbf{MGFM-I} & \textbf{TS-D}
      & \makecell[c]{\textbf{Character}\\\textbf{Zero-shot}}
      & \makecell[c]{\textbf{Radical}\\\textbf{Zero-shot}} \\
    \midrule
      &     &     &     &   22.15  & 11.15  \\
    $\checkmark$ &     &     &     &   31.77  & 13.99  \\
    $\checkmark$ & $\checkmark$ &     &     &   43.59  & 30.14  \\
    $\checkmark$ & $\checkmark$ & $\checkmark$ &     &   45.10  & 30.57  \\
    $\checkmark$ & $\checkmark$ & $\checkmark$ & $\checkmark$
      & \textbf{45.34} & \textbf{31.13} \\
    \bottomrule
  \end{tabularx}
  \caption{Ablation studies of designs in our proposed framework.}
  \label{tab:ablat_all}
\end{table}

\subsection{Ablation Study}
In this section, we perform ablation studies on the handwritten dataset under the character zero-shot setting ($m=500$) and the radical zero-shot setting ($n=50$). Specifically, we evaluate the contributions of each key design in our framework and perform an ablation study on the hyperparameters $\alpha$ and $\beta$.

\textbf{Ablation of Designs in Our Framework.} As shown in Table~\ref{tab:ablat_all}, we observe consistent performance improvements with the addition of each design.
It is noteworthy that MGFM-T improves performance by 9.62\% and 2.84\% in the character and radical zero-shot settings, respectively, demonstrating its effectiveness to exploit the relationships between radicals and strokes to construct a more robust representation space. 
In addition, FCC-M improves performance by 11.82\% and 16.15\% in the character and radical zero-shot settings, respectively. This indicates that fine-grained component matching is more effective than relying solely on holistic representations, as it allows the model to better capture the hierarchical semantics of Chinese characters.

 \begin{figure}[tbp]
  \centering
  \includegraphics[width=\linewidth]{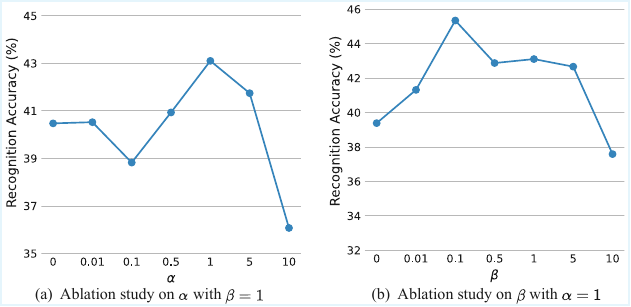}
  \caption{The ablation study on $\alpha$ and $\beta$ under the character zero-shot setting. }
  \label{fig:alpha_beta}
\end{figure}

\textbf{Ablation Study on \(\alpha\) and \(\beta\).}
We study the effect of $\alpha$ and $\beta$ in Eq.~\ref{eq:mul_loss}. The two hyperparameters are used to control the alignment at the levels of stroke/refined stroke and radical/refined radical. We perform ablation studies on the handwritten dataset under the character zero-shot setting ($m=500$). 
Firstly, we fix $\beta = 1$ and vary $\alpha \in \{0, 0.01, 0.1, 0.5, 1, 5, 10\}$ in Figure~\ref{fig:alpha_beta}(a).
When $\alpha = 0$, stroke representations are excluded entirely, and the performance is inferior.
Since stroke representations are inherently low-level and noisy, small values of $\alpha$ (e.g., $0.01$ or $0.1$) may introduce low-quality stroke representations, resulting in reduced performance. 
The model performs the best at $\alpha = 1$, when the model is able to learn decent stroke representations.
These stroke representations can enhance both the stroke and radical representations through our proposed MGFM, ultimately leading to better recognition performance.
When $\alpha$ is set to $5$ or $10$, the excessively high stroke weight may diminish the influence of radical guidance on stroke refinement, resulting in performance degradation. 

Then, we fix $\alpha = 1$ and vary $\beta \in \{0, 0.01, 0.1, 0.5, 1, 5, 10\}$. The results shown in Figure~\ref{fig:alpha_beta}(b) indicate that the model achieves optimal performance when $\beta = 0.1$, suggesting that the dominance of stroke representations leads to the best performance. Fine-grained stroke representations are more expressive for Chinese character recognition than radical-level representations. Nevertheless, radical representations remain important for guiding the model in capturing the architectural decomposition of characters.

\subsection{Visualization of Multi-Granularity Fine-Grained Alignment}
\noindent \textbf{Visualization Method.} To illustrate the interpretability of our model, we visualize the similarity between image tokens and text representations. 
Fine-grained decoupled image-text similarity 
is computed based on the element-wise similarities between image and text tokens. Specifically, for each Chinese character, we visualize the token-level image-text similarity at the levels of stroke, radical, refined stroke, refined radical, and structure.

\begin{figure}[tbp]
  \centering
  \includegraphics[width=\linewidth]{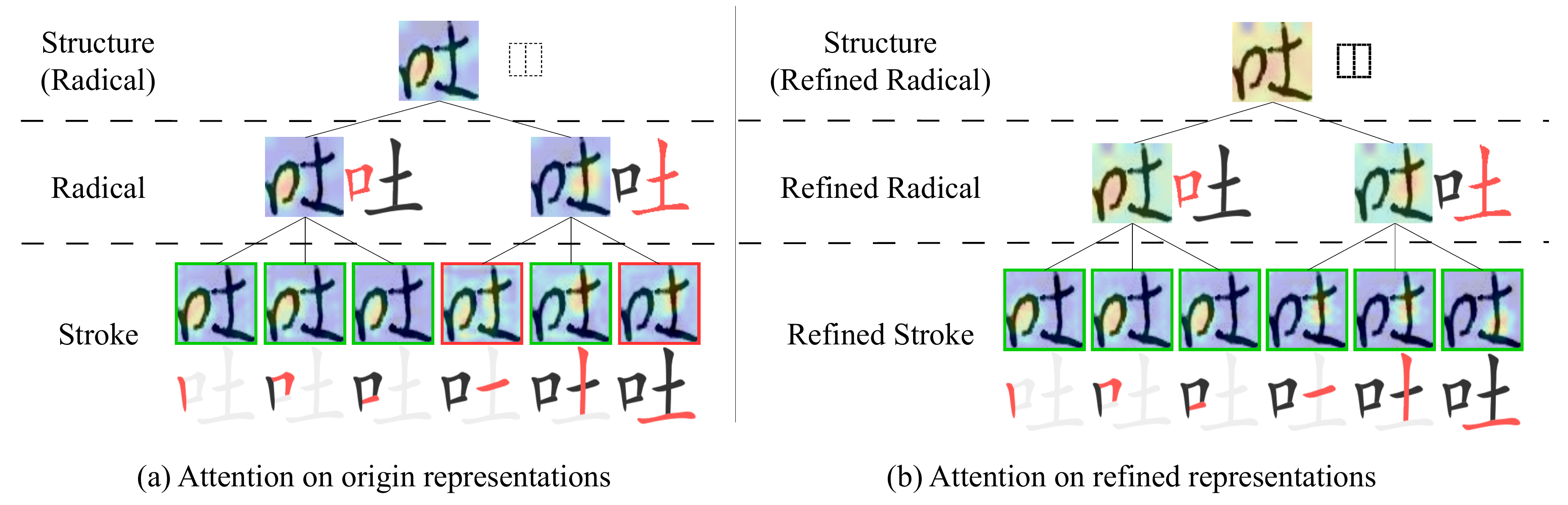}
  \caption{The comparison of attention maps between origin and refined representations. }
  \label{fig:visualize}
\end{figure}

\noindent \textbf{Observations.} 
(1) As shown in Figure~\ref{fig:visualize}, structure representations attend to broader regions than stroke and radical representations, indicating that they capture higher-level layout information rather than the fine-grained component details encoded at lower levels.
(2) Refined stroke representations outperform original strokes in localization. While the original stroke representation misaligns strokes from different radicals (e.g., the 4th stroke in Figure~\ref{fig:visualize}(a)), the refined version correctly attends to the corresponding region (Figure~\ref{fig:visualize}(b)), demonstrating the model’s ability to infer radical affiliation and improve structural awareness.
(3) Refined strokes enable intra-radical discrimination. By leveraging relationships between strokes and their associated radicals, the model can distinguish between structurally similar strokes within the same radical (e.g., final stroke in Figure~\ref{fig:visualize}). This highlights the strength of refined strokes in preserving spatial coherence.

\section{Discussion and Conclusion}
\label{sec:conlcusion}
In this paper, we build upon CLIP and propose Hi-GITA, a Hierarchical Multi-Granularity Image-Text Aligning framework for zero-shot Chinese character recognition. Hi-GITA aims to enhance the model’s ability to capture semantic representations at multiple levels of granularity.
On the image end, we introduce the Image Multi-Granularity Encoder to learn hierarchical representations of Chinese characters. On the text side, we introduce the Text Multi-Granularity Encoder to incorporate both stroke-level and radical-level components to extract multi-granularity representations. In order to better take advantage of the relationships between strokes and radicals, we introduce Multi-Granularity Fusion Modules on both the image and text sides, respectively.
After that, we propose a Fine-Grained Decoupled Image-Text Contrastive loss for effective alignment across semantic levels. Extensive experiments on multiple datasets demonstrate that Hi-GITA significantly improves zero-shot Chinese character recognition performance.

\textbf{Future Work and Limitations.}
Our current approach relies on pre-decoupled stroke and radical sequences. Future work can explore how to directly incorporate the tree-structured decomposition of Chinese characters into the text encoder.
In this study, we focus on Chinese datasets. Extending the method to other East Asian scripts, such as Japanese and Korean, would be an interesting direction for future research. It would also be valuable to investigate the applicability of our method to ancient scripts such as the oracle bone script.

{
    \small
    \bibliographystyle{unsrt}
    \bibliography{main}
}

\newpage
\appendix

\section{Preliminaries}\label{appendix:preliminaries}

\begin{figure}[h]
  \centering
   \includegraphics[width=\linewidth]{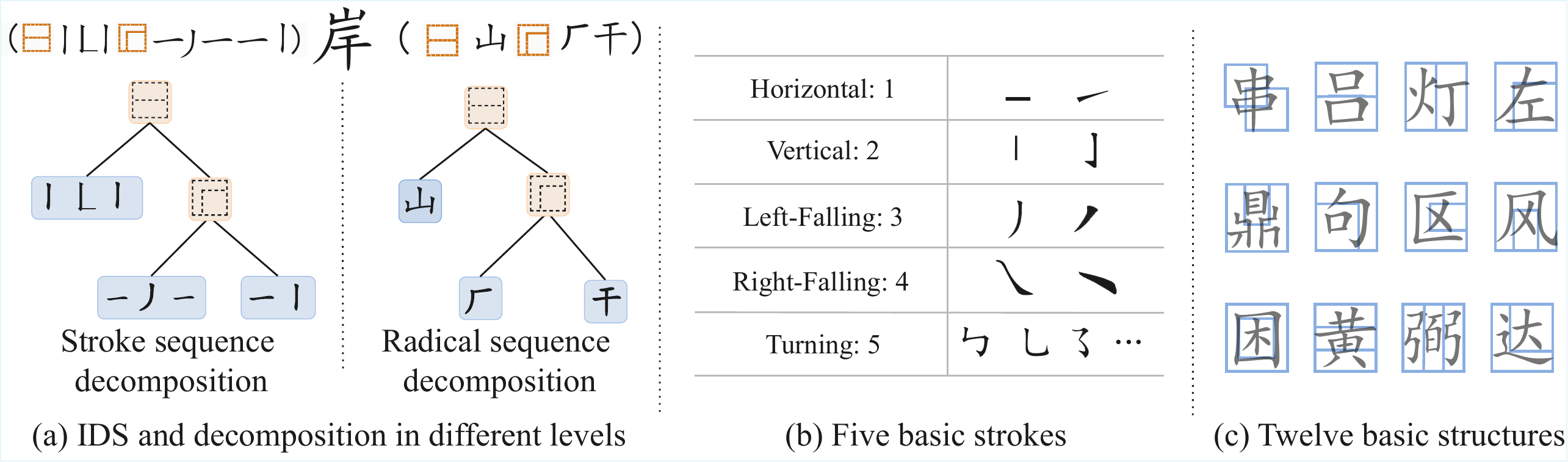}

   \caption{Two types of decomposition for Chinese characters.}
   \label{fig:preliminary}
\end{figure}

Chinese characters are ideographs with a unique internal spatial structure and can be decomposed into a radical or stroke sequence in order, as shown in Figure~\ref{fig:preliminary}(a). According to the Chinese national standard GB18030-2005 \footnote{\url{https://zh.wikipedia.org/wiki/GB_18030}}, there are 70,244 characters, of which 3,755 are commonly used, and the rest are rarely used in daily life. 

\label{sec:pre}

\textbf{Stroke.} Among the components of Chinese characters, strokes are the most fine-grained components, serving as the basis for forming radicals and complete characters. There are five strokes: Horizontal, Vertical, Left-Falling, Right-Falling, and Turning, which are labeled from 1 to 5, as shown in Figure~\ref{fig:preliminary}(b). Each category of stroke contains several instances.

\textbf{Radical.} As fundamental components of Chinese characters, radicals carry semantic information and are composed of finer-grained stroke elements. Compared to the  70,244 Chinese characters, there are only 401 radicals, which greatly reduces the number of classes. 

\textbf{Structure.}
Furthermore, radicals can compose different characters through 12 spatial structures as depicted in Figure~\ref{fig:preliminary}(c), namely overlaid, bove-to-below, left-to-right, upper-left-surround, below-surround, upper-right-surround, left-surround, above-surround, full-surround, above-middle-below, left-middle-right and lower-left-surround.

\textbf{Radical/Stroke Sequence.}
Both stroke and radical sequences contain two types of components: detailed description component and structure component. As illustrated in the radical/stroke decomposition trees in Figure~\ref{fig:preliminary}(a), the leaf nodes correspond to the detailed description component, while the non-leaf nodes represent the structure component.

\section{Closed-Set Recognition Results}
\label{appendix:closed-set}
To assess whether our method generalizes beyond the zero-shot scenarios, we conduct closed recognition experiments on the HWDB1.1~\cite{HWDB} handwritten dataset. All methods are trained/evaluated using the official training/testing split. Table~\ref{tab:close_set_res} shows that our method outperforms the previous state-of-the-art character-, stroke-, radical- and hybrid-based methods by 0.83\%, 3.43\%, 0.36\% and 1.35\%, respectively. This confirms that modeling the correlation between strokes and radicals benefits closed-set recognition performance, beyond the zero-shot setting. The results show the effectiveness and robustness of our proposed method.
\label{Closed-set recognition results}
\begin{table}[t]
  \centering
  \small
  \renewcommand{\arraystretch}{1.0}
  \begin{tabular*}{\linewidth}{@{\extracolsep{\fill}}lcccr}
    \toprule
    Methods & \makecell[c]{Recognition\\ ways} & \makecell[c]{Training \\components}  &\makecell[c]{Inference \\components} & HWDB (\%) \\
    \midrule
    DenseNet~\cite{huang2017densely}    & Classification & C  & C             & 94.32 \\
    ViT-L~\cite{dosovitskiy2020image}    & Classification & C & C               & 95.13 \\
    DirectMap~\cite{zhang2017online}     & Classification & C & C               & 96.25 \\
    DropSample-DCNN~\cite{yang2016dropsample} & Classification & C & C             & 96.57 \\
    \midrule
    SD~\cite{chen2021zeroshot}           & Auto-regressive & S   & S            & 93.97 \\
    \midrule
    RAN~\cite{zhang2020radical}          & Auto-regressive & U \& R & U \& R          & 92.28 \\
    HDE~\cite{CAO2020107488}             & Auto-regressive & U \& R & U \& R         & 95.63 \\

    RZCR~\cite{diao2022rzcr}             & Auto-regressive & U \& R & U \& R         & 96.45 \\
    UCR~\cite{li2025ucr}                 & Auto-regressive & U \& R  &U \& R         & 96.95 \\
    CCR-CLIP~\cite{CCR-CLIP}             & Retrieval & U \& R &U \& R         & 97.04 \\
    \midrule
    STAR~\cite{zeng2023zero}             & Auto-regressive & U \& R \& S  & S & 94.24 \\
    RSST~\cite{yu2023chinese}            & Auto-regressive & U \& R \& S  & U \& S & 96.05 \\
    \midrule
    Ours                                & Retrieval & U \& R \& S & U \& R \& S  & \textbf{97.40} \\
    \bottomrule
  \end{tabular*}
  \caption{Comparison to existing methods on the closed-set recognition setting. ``Recognition ways'' represents which recognition way of the method is based on. ``Training components'' and ``Inference components'' represent which character components are used during training and inference, respectively. ``U'', ``R'', and ``S'' denote structure component, radical component, and stroke component, respectively.}
  \label{tab:close_set_res}
\end{table}

\begin{figure}[tbp]
  \centering
  \includegraphics[width=\linewidth]{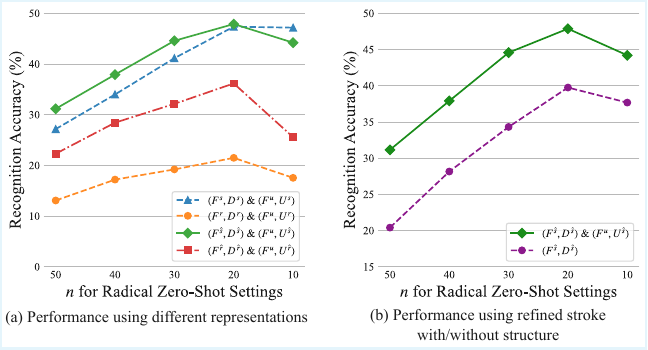}
  \caption{Performance comparison of different representations used for inference in the radical zero-shot settings. $\left( \cdot\right)$ represents a pair of image representations and their text counterparts used for retrieval.}
  \label{fig:test_diff_rep}
\end{figure}

\section{Inference with Different Representations}
\label{appendix:RZS-diff_rep}
To more comprehensively compare the performance of different representations, we also conduct experiments on the handwritten dataset under the radical zero-shot setting. We first evaluate the performance by matching both detailed description and structure components from four types of text sequence representations with their corresponding image representations. The results are presented in Figure~\ref{fig:test_diff_rep}(a). As in the character zero-shot setting, refined stroke representations generally yield the best results.
It is worth noting that different levels of representations may perform differently between the character and radical zero-shot settings. Specifically, in the radical zero-shot setting, the performance ranking of different semantic representations becomes: refined stroke, stroke, refined radical, and radical. This may be attributed to the degradation of radical representations for unseen radicals in this setting. Since there are rarely zero-shot problems for strokes, stroke representations are more reliable. As a result, strokes outperform radicals in this setting. Benefiting from the semantic interaction between strokes and radicals, refined strokes outperform original strokes and demonstrate robust performance across both character and radical zero-shot settings.

Then, we evaluate the performance using refined stroke with/without structure for recognition. As shown in Figure~\ref{fig:test_diff_rep}(b), similarly to the results in the character zero-shot setting, removing the structure component results in an average performance drop of 9.08\% in the radical zero-shot setting, while removing the detailed description component yields only 0.57\% recognition accuracy.

\section{Comparison of Recognition Results with CCR-CLIP}
 \begin{figure}[tbp]
  \centering
  \includegraphics[width=\linewidth]{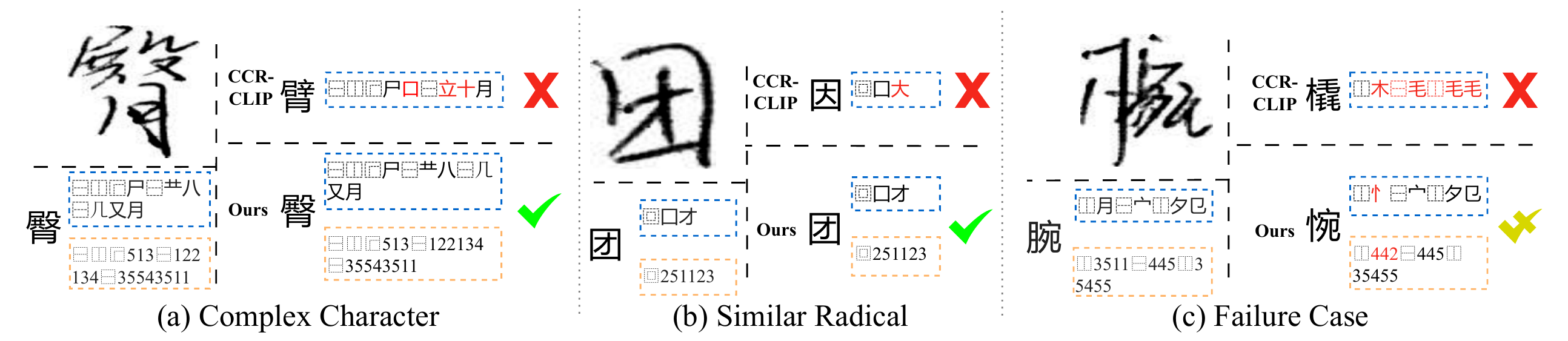}
  \caption{Comparison of recognition results between CCR-CLIP and our method. In each subgraph, the upper left, lower left, upper right, and lower right represent the original image, the ground truth, recognition result of CCR-CLIP, and recognition result of our method, respectively. Blue dotted boxes highlight the radical sequences, while orange dotted boxes indicate stroke sequences. Incorrect recognition results are indicated in red.
  }
  \label{fig:compared_CCR_CLIP}
\end{figure}

To validate the effectiveness of hierarchical representations, we visualize and compare the recognition results between our method and the state-of-the-art approach CCR-CLIP~\cite{CCR-CLIP}. As shown in Figure~\ref{fig:compared_CCR_CLIP}(a), for the character ``臀'', CCR-CLIP incorrectly predicts ``臂'', while our method predicts correctly. This suggests that, for characters with complex radicals, relying solely on radical-level information is insufficient for accurate recognition. 
As shown in Figure~\ref{fig:compared_CCR_CLIP}(b), for the character ``团'', the radicals ``大'' and ``才'' may look similar in the handwritten form. However, fine-grained stroke-level representations can reveal clear structural differences between them. Specifically, ``大'' is composed of a horizontal stroke, a left-falling stroke, and a right-falling stroke, while ``才'' consists of a horizontal stroke, a vertical stroke, and a left-falling stroke. Our model effectively leverages these stroke-level distinctions to improve recognition accuracy.

Figure~\ref{fig:compared_CCR_CLIP}(c) also shows a case of complex and severely distorted handwritten character. Although both our method and CCR-CLIP fail to recognize the character ``腕'', our model correctly identifies the right-side radical ``宛''. In contrast, CCR-CLIP fails to extract any meaningful component. This highlights the advantage of leveraging hierarchical representations for robust recognition under challenging conditions.

\end{CJK}

\end{document}